\documentclass[pdflatex,sn-mathphys-num]{sn-jnl}

\usepackage{graphicx}%
\usepackage{multirow}%
\usepackage{amsmath,amssymb,amsfonts}%
\usepackage{amsthm}%
\usepackage{mathrsfs}%
\usepackage[title]{appendix}%
\usepackage{xcolor}%
\usepackage{textcomp}%
\usepackage{manyfoot}%
\usepackage{booktabs}%
\usepackage{algorithm}%
\usepackage{algorithmicx}%
\usepackage{algpseudocode}%
\usepackage{listings}%
\usepackage[T2A]{fontenc}%
\usepackage{threeparttable}%
\usepackage{makecell}
\usepackage{hyperref}
\usepackage{tabularx}
\usepackage{array}
\usepackage{longtable}
\usepackage{ltablex}
\usepackage{xltabular}
\usepackage{caption}

\theoremstyle{thmstyleone}%
\theoremstyle{thmstyletwo}%

\theoremstyle{thmstylethree}%

\newcolumntype{L}{>{\raggedright\arraybackslash}X} 
\newcolumntype{C}{>{\centering\arraybackslash}X}   
\begin{document}

\title[Article Title]{RusLICA: A Russian-Language Platform for Automated Linguistic Inquiry and Category Analysis}




\author[1]{\fnm{Elina} \sur{Sigdel}}\email{essigdel@ipran.ru}

\author[1]{\fnm{Anastasia} \sur{Panfilova}}\email{panfilovaas@ipran.ru}

\affil[1]{\orgname{Laborarory of AI application in Psychology, Institute of Psychology (RAS)}}


\abstract{Defining psycholinguistic characteristics in written texts is a task gaining increasing attention from researchers. One of the most widely used tools in the current field is Linguistic Inquiry and Word Count (LIWC) that originally was developed to analyze English texts and translated into multiple languages. Our approach offers the adaptation of LIWC methodology for the Russian language, considering its grammatical and cultural specificities. The suggested approach comprises 96 categories, integrating syntactic, morphological, lexical, general statistical features, and results of predictions obtained using pre-trained language models (LMs) for text analysis. Rather than applying direct translation to existing thesauri, we built the dictionary specifically for the Russian language based on the content from several lexicographic resources, semantic dictionaries and corpora. The paper describes the process of mapping lemmas to 42 psycholinguistic categories and the implementation of the analyzer as part of RusLICA web service.}

\maketitle

\section{Introduction}\label{sec1}

In recent years, research interest in analyzing human behaviour, social dynamics, and personal traits through written texts in both online and offline domains, has significantly grown. The increasing number of textual data from social media, linguistic corpora, news articles, forums, personal blogs, transcribed content from the colloquial speech, etc. require specialized automated tools capable of capturing significant psychological and linguistic information from texts. 

One of the most used tools in the current field is the Linguistic Inquiry and Word Count (LIWC) framework, which was originally developed in the 1990s \cite{pennebaker1999linguistic}. The last version of LIWC provides more than 100 categories \cite{boyd2022development} referring to the various psychological states and processes. Since LIWC was developed, it has been widely used for detection of psychological states \cite{holtzman2019linguistic}, personal traits \cite{koutsoumpis2022kernel}, social processes \cite{kwon2018will} and communication analysis, the assessment in clinical psychology \cite{huston2019exploratory}, and  the analysis of other aspects of human behaviour. The main principle the approach follows is computation of word occurrences belonging to significant psychological categories, which makes LIWC a  closed-dictionary approach \cite{kern2016gaining}. Therefore, LIWC can be a well-suited and highly reliable tool for psychological studies for its empirical validation of the authors, intuitive and flexible software.

Despite LIWC's broad applicability, the original service was developed to conduct analysis of English texts. Although some of the latest versions \cite{pennebaker1999linguistic, boyd2022development} contained adaptations of the dictionary to other languages, including structurally different languages from English (Arabic, Japanese, Turkish,  Russian, etc.), adaptation for languages with rich morphology or typologically distant from English remains challenging. It is necessary to take into account differences between English and the target language in cultural specificities, word-formation, and grammatical structure, as they may affect the representations of psychological constructs in texts.

Common approaches to thesaurus adaptation involve applying manual translation or machine translation with subsequent validation when it is deemed necessary. However, direct translation cannot cover all items for semantic fields observed in a dictionary and needs both validation and extension dictionaries with specific lexical content for the target language. Psychometrical validation in developing thesaurus also requires expert involvement with linguistic and psychological expertise, which can be resource-consuming \cite{boyd2022development, panicheva2020matching}. Some empirical studies devoted to evaluating the effectiveness of dictionary adaptations and analyzing cross-linguistic comparisons between them have noted that translation of dictionaries without consideration of linguistic and cultural particularities may lead to biased results \cite{duduau2021performing}.

Early attempts to adapt a thesaurus for the Russian language \cite{kailer2011russian} were conducted without additional validation, resulting in limited application of the Russian version in further research \cite{panicheva2020matching}. One of the recent studies demonstrates the efficiency of using automated tools aggregating semantic information for Russian \cite{panicheva2020matching} to expand coverage of observed semantic fields and improve the quality of analyzing psycholinguistic  categories. 

In addition, as a closed-vocabulary approach, LIWC has been criticized for its limitations as an NLP approach \cite{panicheva2020matching}, since its methodology is based on word counting without consideration of the general context in texts. For languages with a rich morphological system, the current method may also failed to extract psycholinguistic information accurately. In such languages grammatical meaning can be expressed through inflectional morphemes (e.g. suffixes, prefixes, etc.). These grammatical specificities cannot be captured through LIWC.  

Therefore, the current study proposes an approach that computes linguistic features with automated NLP methods and examines lexicon reflecting significant psychological dimensions. To adapt the LIWC methodology for Russian and make the results of analysis more representative and complex for both types of tasks, psychological and linguistic, computations of general statistics, word frequency, lexical, morphological, syntactic characteristics of texts were obtained. We employ automated linguistic parsers and language models to obtain scores for linguistic categories and build a LIWC-like dictionary to analyze lexical categories. The lexical component was constructed using content from semantic dictionaries, enriched with data from Russian National Corpus\footnote{Russian National Corpus \url{  https://ruscorpora.ru/en}} (RNC) and RuWordNet\footnote{RuWordNet \url{  https://ruwordnet.ru/ru/}} thesaurus. The suggested list of lexical categories was designed to be comparable with the categories employed in LIWC \cite{pennebaker2015development, boyd2022development}, at the same time it was expanded to include semantically significant categories for the Russian language.

Consequently, in the current study we developed an automated analyzer for Russian texts. The tool computes values across 96 categories covering different linguistic and psychological aspects of language use, considering linguistic and cultural specificities. Based on the analyzer, we developed the first version of  the publicly available web service, RusLICA (a Russian-language tool for Linguistic Inquiry and Category
Analysis), that analyzes dimensions mentioned above.

\section{Related work}\label{sec2}

LIWC \cite{pennebaker1999linguistic, pennebaker2001linguistic, pennebaker2007linguistic, pennebaker2015development, boyd2022development} framework is widely used in analyzing the occurrence of significant psycholinguistic categories in texts. Based on a predefined thesaurus, it counts the number of lexical entries, and reports a score for a category as the ratio between count of words in a given text belonging to that category and total number of words in a text. The resulting scores for each category are computed as follows \cite{kern2016gaining}: 
\begin{equation}\label{eq:liwc_prob}
P(\text{category}) = \sum_{word \in category} p(word) = \frac{ \sum_{word \in category}count(word) }{ N\_word }
\end{equation}

LIWC was developed primarily for the English language to obtain psychological dimensions represented in a given text considering lexical peculiarities. Since the original service was developed, the list of linguistic and psychological dimensions and their content have been slightly changed through several releases. 

The latest version of LIWC \cite{boyd2022development} includes over 100 dimensions and contains over 12,000 lexical entries (words, stems, emoticons, and phrases) in the internal dictionary. It also provides dictionaries for other languages that were adapted based on the earlier LIWC versions \cite{pennebaker2001linguistic, pennebaker2007linguistic, pennebaker2015development}, such as the Russian translation of the dictionary released in 2007 \cite{kailer2011russian}. 

However, the process of adapting a dictionary built for English to other languages is a challenging task due to structural differences between the target language and cultural specificities in expressing psychological constructs. Earlier studies on LIWC adaptation sought to incorporate these aspects into a dictionary for the target language. The most common adaptation strategies obtain automated translation using machine translation tools with further manual correction \cite{van2017electronic, meier2019liwc}, and employing expert-led manual translation with subsequent expansion and validation by linguists and psychologists \cite{duduau2022development, bjekic2012razvoj}. For languages structurally different from English, especially with rich inflectional and morphological systems, such as Serbian \cite{bjekic2012razvoj} and Romanian \cite{duduau2022development}, adaptation requires a specific approach. To consider these morphological aspects, it is necessary to explicitly include sets with possible derivations and inflections of word forms instead of direct translation lexical entries from the original English dictionaries. The resulting dictionary can be enriched with synonyms. 

For the Russian language, several attempts to adapt the LIWC dictionary were also applied.

The earliest version of adaptation involves a direct translation of the English LIWC dictionary released in 2007 into Russian \cite{kailer2011russian}. However, as other researchers have noted, the results were not subsequently validated for internal consistency. As a result, the current dictionary was rarely used in analyzing Russian texts for psycholinguistic categories \cite{panicheva2020matching}. A subsequent study using this adaptation \cite{litvinova2018dynamics} demonstrated the tool’s utility in analyzing a large corpus of texts related to social processes such as suicide, and its capability to extract both stable and varying author style characteristics. It was noted that the internal dictionary needs to be extended and validated by experts to make the psychological measurements reliable for further research. The analysis revealed that categories referring to cognitive processes and space were highly stable, while personal and overall pronouns, emotions, negations, and time were moderately stable \cite{panicheva2020matching, litvinova2018dynamics}.

Hence, one of the recent studies demonstrates a strategy for adaptation of the LIWC dictionary for the Russian language including expansion of content for categories and validation to capture significant psychological and linguistic aspects of texts more accurately. The authors adapted a dictionary for 8 categories, matched them to the Russian thesauri, and conducted several experiments using the resulting dictionary \cite{panicheva2020matching}. Dictionaries were constructed using machine translation and subsequent semi-automated extension of the thesauri for the following 8 categories: Cognitive, Social, Biological processes, Perceptual processes (See, Feel, Hear). To extend the thesaurus, synonym dictionary and RuWordNet thesaurus (based on RuThes\footnote{RuThes \url{  https://www.labinform.ru/pub/ruthes/}}) were applied to develop more accurate adaptation. As a result, authors offer the validated version of the Russian dictionary. Their study demonstrates that dictionaries for English and Russian could differ significantly, while results for author profiling tasks obtained on Russian texts remain reasonable. Moreover, based on the modest matches between the translated lexical entries of original LIWC and selected content from Russian thesauri, a simple direct transfer of the content without extension and validation remains insufficient. Authors also argue that the content and structure of the original LIWC dictionary could be subjective, which may influence the results of adaptation \cite{panicheva2020matching}.

\section{Categories for Text Analysis}\label{sec3}

The current study proposes a set of 96 categories for text analysis considering linguistic and cultural specificities of the Russian language. All categories are organized hierarchically,  reflecting the method used for their extraction. Groups and examples of corresponding categories are presented in Table~\ref{tabgeneral}.

{
\footnotesize 
\setlength{\tabcolsep}{5pt}
\begin{xltabular}{\textwidth}{@{} L  C C C @{}}
\caption{Groups of categories computed by service}\label{tabgeneral}\\
\toprule
\makecell{Group} & \makecell{Example of\\categories} &  \makecell{Number of\\categories} & \makecell{Approach of features\\ extraction / Sources}\\
\midrule
General statistical features    & Number of words, Number of repeated words, IPM frequency, etc.   & 15 & Frequency dictionary  \\
\\
Dependency-based syntactic features   & Average syntactic tree depth, Number of adnominal clauses, etc.   & 11  & SpaCy, \texttt{‘ru\_core\_news\_lg’} model  \\
\\
Morphological features    & Number of 1st person pronouns, Number of adjectives, Number of verbs in the passive voice, etc.   & 27  & SpaCy, \texttt{‘ru\_core\_news\_lg’} model  \\
\\
Lexical features representing psycholinguistic dimensions  & Swear words, Family, Fear, Love, Motion, etc.   & 42  & Semantic dictionaries, 
RNC,
RuWordNet
  \\
\\
Classification outputs  & Emotion detection (7 classes) & 1  & Pretrained models:
\textit{‘Aniemore/rubert-tiny2-russian-emotion-detection’}  \\

\botrule
\end{xltabular}
}

To compute scores for observed categories, texts were preprocessed through the following pipeline: conversion of a given text to lowercase, replacement of hashtags, numbers, URLs, and emojis with generalized special tokens (\texttt{‘HASHTAG’}, \texttt{‘NUM’}, \texttt{‘URL’}, \texttt{‘EMOJI’}), removal of duplicated punctuation. This preprocessing was applied to a text prior to feature (i.e., scores for categories) extraction. The  \ref{sec11} section provides details about the resulting pipeline and technical specifications.

A comprehensive explanation of the full list of features the service contains is provided in Appendix~\ref{allcategories} (for the non-lexical categories) and in Appendix~\ref{secA1} (for hierarchical structure of lexical categories).

\subsection{Statistical and frequency categories}\label{subsec3}

The current group comprises categories (i.e., features) that capture common statistical properties of a given text, such as word count, average word length, the number of repeated words, and lexical frequency measurements.

Computation of statistical features was conducted using SpaCy\footnote{SpaCy \url{ https://spacy.io/}} with the Russian-language model  \texttt{ru\_core\_news\_lg} for tokenization and lemmatization. Based on SpaCy results, values for the number of tokens per text and per sentence were calculated. To analyze the number of repeated words and the number of appearances of the most repeated word, a frequency dictionary of lemmas that appeared more than once in a given text was created. Special tokens mentioned above were included in the computation procedure for statistical categories, except analyzing repeated words to avoid noise in the results.

Frequency categories were computed  using resources comprising two precomputed frequencies of Russian words: items per million (IPM frequency) and a coefficient reflecting the distribution of a word in different types of texts (D frequency). Values for both IPM and D were extracted from the RNC-based frequency dictionary \cite{lyashevskaya2009dict}. Values from the dictionary were mapped to the word lemmas in a given text considering Part-of-Speech tags. To ensure lemmatization aligned with the source dictionary, MyStem\footnote{MyStem \url{ https://tech.yandex.ru/mystem}} was applied, as it resolves the  part-of-speech homonymy for each word in a sequence. Values of IPM and D were extracted from the frequency dictionary only for matching lemmas with corresponding part-of-speech tags in a given text. If a text did not contain lemmas (with their POS tags) that were represented in the dictionary, the resulting values for categories were set to -1.

\subsection{Syntactic categories}\label{subsec4}

Syntactic categories (i.e., features) are computed based on the dependency structure. Thus, dependency trees for each sentence in a given text were built using SpaCy with the \texttt{ru\_core\_news\_lg} model. 

We chose potentially significant syntactic and morphological features, which could be derived from the parsing of dependency tree results, as well as based on earlier study analyzing the characteristics of Russian-speaking students’ essays \cite{vinogradova2017multi}.  For each token, a value of its dependency tag was obtained labeling the corresponding syntactic relation from the Universal dependencies list\footnote{Universal Dependency Relations \url{ https://universaldependencies.org/u/dep/}}. Thus, the analyzer computed counts of various dependency relations (i.e., UD relations types), such as conjuncts (\texttt{‘conj’} tag), discourse elements (\texttt{‘discourse’} tag), etc. The analyzer also calculated the depth of the dependency tree for each sentence in a given text.

 The full list of syntactic categories is described in Appendix
~\ref{allcategories}.

\subsection{Morphological categories}\label{subsec5}

All features that represent morphological characteristics of a given text were obtained using a parser and model for the Russian language (the same model that was used for syntactic categories) from SpaCy. Based on parsing results, the value of the corresponding Part-of-Speech tag from the Universal POS\footnote{Universal Feauteres with POS tags and inflectional features \url{  https://universaldependencies.org/u/feat/}} list for each token was extracted. Due to the rich inflectional system of Russian, detailed morphological information for tokens was also extracted to compute features scores.

Therefore, the number of coordinating, subordinate conjunctions, pronouns of the 1st, 2nd and 3rd persons, adjectives (including degrees), adverbs, and verbs (annotated for aspect, voice, tense, and mood) were counted. The count of verbs (including the morphological attributes) functioning as a root of the syntactic tree from each sentence in a given text was computed separately.

The full list and description of the morphological categories could be found in Appendix~\ref{allcategories}.

\subsection{Lexical categories}\label{subsec6}

The current version of the analyzer provides 42 categories reflecting the lexical characteristics of a given text. As recent studies focused on several categories to demonstrate how a thesaurus can be adapted to the Russian language \cite{panicheva2020matching}, we examined certain categories from LIWC and developed our dictionary accordingly. 

We used the following categories from the latest version of LIWC \cite{boyd2022development} to build a particular hierarchical structure of our dictionary: Negations, Causation, Swear words, Motion, Time, Visual perception, Auditory perception, Space, Death, Sexual, Mental health, Religion, Health, Female references, Male references, Family, Friends, Anger,  Social behavior, Communication, Sadness, All-or-none (as an absolutist language), Certitude, and Cognition.

The resulting dictionary comprises 7492 unique lemmas. Nevertheless, it is essential to acknowledge that some lemmas are included in several categories. For instance, \textit{‘внучка’}  (granddaughter) occurs in both female referent and family. 

Thus, the total number of lexical entries across categories is 8309. 

\subsubsection{Data collection}\label{subsubsec7}

Instead of directly translating the English LIWC dictionary using automated methods, we constructed the core of the dictionary using semantic dictionaries and expanded the set of words using data from RNC and RuThes thesaurus adapted to the WordNet format. 

Dictionaries that organize words and phrases according to their meanings and semantic relations, and thesauri with relevant terms that have already been verified by experts were used to develop a dictionary of the current service. Semantic dictionaries contain sets of words including their inflectional forms (e.g., various forms of a word with prefixes, suffixes belonging to a particular semantic field). Consequently, applying content from dictionaries enables the identification of the majority of words and their derivatives within a semantic category.

To construct the content for lexical categories the following dictionaries were used: 

\begin{enumerate}[1.]

\item Based on the study analyzing the lexical representation of emotions in the Russian language \cite{babenko2021emotions}, the categories referring to emotion domains were built. The current dictionary systematically describes emotive vocabulary, highlighting the conceptual and semantic features of words for each emotion. Moreover, it contains not only words that explicitly mark emotion, but also a set of words describing the external manifestation of emotions, characteristics of an emotional state, words with emotional connotation. It significantly increases the set of emotions that can be extracted from a given text. Unlike LIWC, which categorizes emotions into positive and negative, the current study attempts to identify different types of emotions specifically for the Russian language. As a result, categories \textit{'вера'}, \textit{'счастье'}, \textit{'радость'}, \textit{'любовь'}, \textit{'вдохновение'}, \textit{'горе'}, \textit{'грусть'}, \textit{'беспокойство'}, \textit{'страх'}, \textit{'одиночество'}, \textit{'стыд'}, \textit{'вражда'}, \textit{'высокомерие'}, \textit{'обида'}, \textit{'любопытство'}, \textit{'влечение'}, \textit{'сомнение'} were built.

\item The dictionary hierarchically organizes Russian verbs by lexico-semantic fields and subfields distinguished by categorical-lexical semantics \cite{babenko1999verbs}. The dictionary provides inflectional versions (perfective/imperfective, transitive/intransitive, reflexive/non-reflexive) of verbs that are grouped around the current semantic fields. As verbs can affect the semantics and pragmatics of the other dependent words in a sentence and denote actions, states, processes, and relationships, analyzing them separately is important to collect content for some categories. We used the dictionary to extract words for movement, personal and social relations, speech and conversation, etc.

\item Another semantic dictionary systematizing Russian words by groups contains about 300 000 words and phrases \cite{shvedova1998russkiy}. We used Volume 3 describing the semantics of abstract nouns. Semantics of the abstract nouns reflect the mental sphere of a person, the self-representation and the representation of the world, state of being, relations, etc. \cite{shvedova1998russkiy}. Therefore, employing comparable abstract nouns could represent an individual's cognitive processes.  We extracted fields related to death, cognition, religion, modality, physical traumas and illnesses, mental health, birth and sexuality, time, space, general representations of motion, speech. The selection of categories is close to the corresponding LIWC categories; however the content may differ from the English dictionary as it represents the characteristics of the Russian language.

\item The intensifiers for the Russian language were selected from the dictionary constructed from Dostoyevsky’s letters\footnote{Sharapova Ye.V. \url{ https://ruslang.ru/intens_Dostoevsky}}. Intensifiers denote a high degree of a graded feature of words or entire phrases (e.g. \textit{ 'слишком'}, \textit{ 'совершенно”'}, \textit{ 'отчаянно'}). The dictionary was built on the diachronic analysis of  texts from 1832 to 1881 and some words could be out of scope in texts written in social media. Nevertheless, the set of words could remain efficient enough for the observed field. 

\end{enumerate}

The process of word selection involved several steps. First of all, words referred to the categories were extracted from the observed dictionaries. Some words were excluded from lists after sets of the given categories were composed. Removing words helped us to minimize the imbalance between categories, as words can be used in a large number of contexts, and the observed category may not be the \textbf{main} for them. For verification, information from the RNC for chosen words that could have an ambiguity was used to determine possible contexts. In cases, where words tended more towards a category outside of the significant fields, they were excluded from the list. The analysis was carried out by introspection. 

The corpus approach was also applied to extract the words for some categories. For instance, swear words, additional words related to time and space, denotations for female and male referents were extracted directly from RNC. The contexts for words were examined in a similar manner to words from the semantic dictionaries. 

However, using the approaches described above could not guarantee comprehensive coverage of the observed lexico-semantic fields. The recent study \cite{panicheva2020matching} demonstrated that using automated methods such as RuWordNet to expand the word lists for LIWC could be effective and reflect more peculiarities of the Russian language in the analysis of the lexical characteristics than the results of simple machine translation approach. 

Therefore, the set of words was expanded using the RuWordNet thesaurus to make the dictionary for the service enhanced. RuWordNet was built on the RuThes thesaurus adapted to the WordNet format. Originally, RuThes \cite{lukashevich2010thesauri, loukachevitch2014ruthes} is a hierarchical network containing 31 thousand language concepts and more than 111 thousand relations between them. As RuWordNet is a thesaurus that consists of synsets for three parts of speech (adjectives, nouns, and verbs) and defines a set of relations between synsets, we used the following relations to expand the list of words: hyponym-hyperonym, meronym-holonym, subject area (domain), synonyms from different parts of speech. Thus, words from the corresponding domains or synsets with types of relation mentioned above were collected to extend the set of some categories. Then some words were removed from the dictionary after examination of possible contexts in RNC and introspection.

\subsubsection{Dictionary}\label{subsubsec8}

The resulting version of the built dictionary includes categories that were collected based on the structure of original LIWC-22 \cite{boyd2022development}. However, some categories taken from LIWC were transformed and expanded. Thus, the content of the categories may differ from the English LIWC version. For example, instead of focusing on past, present and future as subcategories within Time, we retained a Time category and extracted time of verbs as morphological categories in addition. Moreover, instead of using subcategories Family and Friends as social referents, the content for them was expanded to cover corresponding semantic fields and settled as Family and Friendship. 

Categories in the obtained dictionary are organized hierarchically. We ordered lexical categories in the following manner.

\textbf{Linguistic Dimensions}: this group contains categories that relate to the grammatical aspects and form of utterance, rather than to its thematic content. Analyzing the chosen categories enables estimation of the degree of author’s confidence (Modality), emotional intensity (Intensifiers) and reflects specific parts of the syntactic structure (Negations).   

\textbf{Psychological processes}: this group was organized considering the structure that was embedded in the original LIWC \cite{boyd2022development}. It comprises such groups of categories as Affective processes, Social processes,  Physical, Perception, Motives, and Cognition that could reflect the speaker’s mental activity and concerns about the world, which could be verbalized. Generally, categories included in this dimension are meaningfully relevant to psychology.  

\textbf{Lifestyle}: at the current moment this dimension consists of  Metaphysical processes group (Religion). The category was assigned to the Lifestyle dimension as it reflects sustainable practices, cultural identity, and worldview aspects of a speaker’s daily life.

\textbf{Time and Space orientation}: this group comprises categories related to time and space. Emphasizing them as a separate dimension could help in analyzing the cognitive orientation of a speaker, taking into account the events, objects and their position in the world and timeline. Analyzing given categories could also help to define a speaker's experience in the past or feeling distanced. 

Detailed information of structure of the lexical categories and subcategories is presented in Appendix~\ref{secA1}. 

Before applying computations to a given text, all words in the dictionary were normalized with MyStem to make the following assessment procedure more accurate. Unlike the original LIWC \cite{boyd2022development} that contains words, stems, emoticons and phrases in the internal dictionary, the current dictionary consists of lemmas for single words. Therefore, a normalized version of the dictionary is used as a resulting version applied by the analyzer.

\subsubsection{Scoring methodology}\label{subsubsec9}

To compute scores for the lexical categories accurately, lemmatization with MyStem was applied as a first step to a given text. This method resolves POS ambiguity. As the dictionary was lemmatized using the same approach, lemmas from a text are matched with lemmas from the dictionary. 

Thus, the percentage of words belonging to each category was computed according to the original LIWC computing technique.

\subsection{Classification on pretrained models}\label{subsec10}

The analyzer also supports classification using pretrained language models as an additional source of text analysis.

The current version of the service uses a pretrained model \textit{Aniemore/rubert-tiny2-russian-emotion-detection} from Hugging Face platform \cite{davidchuk2023animore} to determine the dominant emotion of a text. 

The model was trained on the CEDR dataset. Originally, the corpus contained 9410 texts in Russian, annotated with five categories of emotions. The data was collected from Russian-language resources, including posts from social networks (\texttt{LiveJournal}), news resources (\texttt{Lenta.ru}) and microblogs (\texttt{Twitter}).

The authors of the model extended the set of emotions. As a result, a fine-tuned version of \textit{rubert-tiny2} model predicts 7 emotions: neutral, happiness, sadness, enthusiasm, fear, anger, and  disgust. 

Due to the specifics of the data, a model fine-tuned on this corpus is able to accurately identify emotions in short texts. It can perform worse with long texts or specific domains, such as texts with narrowly focused specialized vocabulary.

\section{Service}\label{sec11}

As one of the main goals of the current study is to make the analysis of significant linguistic and psycholinguistic features in Russian texts more accessible for researchers, we developed a free-to-use web application RusLICA (\href{https://ruslica.ipran.ru/}{ruslica.ipran.ru}). The suggested service combines a linguistic analyzer with an integrated methodology of LIWC to compute the obtained categories described earlier.

RusLICA provides researchers an opportunity to analyze a large corpus of texts by uploading a dataset containing written texts in Russian. It computes scores for each of 96 categories. 

The full functionality of the service is available after registration.

\subsection{System architecture and technical requirements}\label{subsec12}

The service was designed to accept and process written Russian texts with original punctuation. Currently, users can upload machine-readable files to the platform in several formats: Comma Separated Values (\texttt{.csv}) or Microsoft Excel (\texttt{.xlsx}). Target texts should be stored in a column named \texttt{‘text’}. 

The “Analysis” section allows users to upload a selected file and choose between two analysis options: general computation across 96 categories, or a computation applying more detailed analysis to lexical categories. Then the service queues the uploaded file and starts its analysis. As the analysis is completed, resulting files containing computed scores can be downloaded in \texttt{CSV} format with a tab separator. 

For the detailed evaluation the service counts the occurrences of each word from lexical categories in a text and writes the results to a separate file. As a result, a \texttt{JSON} (\texttt{.json}) file containing all texts from a dataset with count number of word occurrences referred to the categories is generated.

Moreover, the current web application provides an opportunity to manage the history of  analysis results obtained for files uploaded to the server. It allows a user to view, download, and delete the results of analysis of all files that have been uploaded to the service.  

The service can handle files of any size. However, the processing time is strictly limited: a file cannot be processed for more than 12 hours. If the analysis takes longer than the time limit, the service provides corresponding information to a user with the termination status.

\subsection{Data processing pipeline}\label{subsec13}

\begin{figure}[h]
\centering
\caption{Data processing pipeline}\label{fig1}
\includegraphics[width=0.9\textwidth]{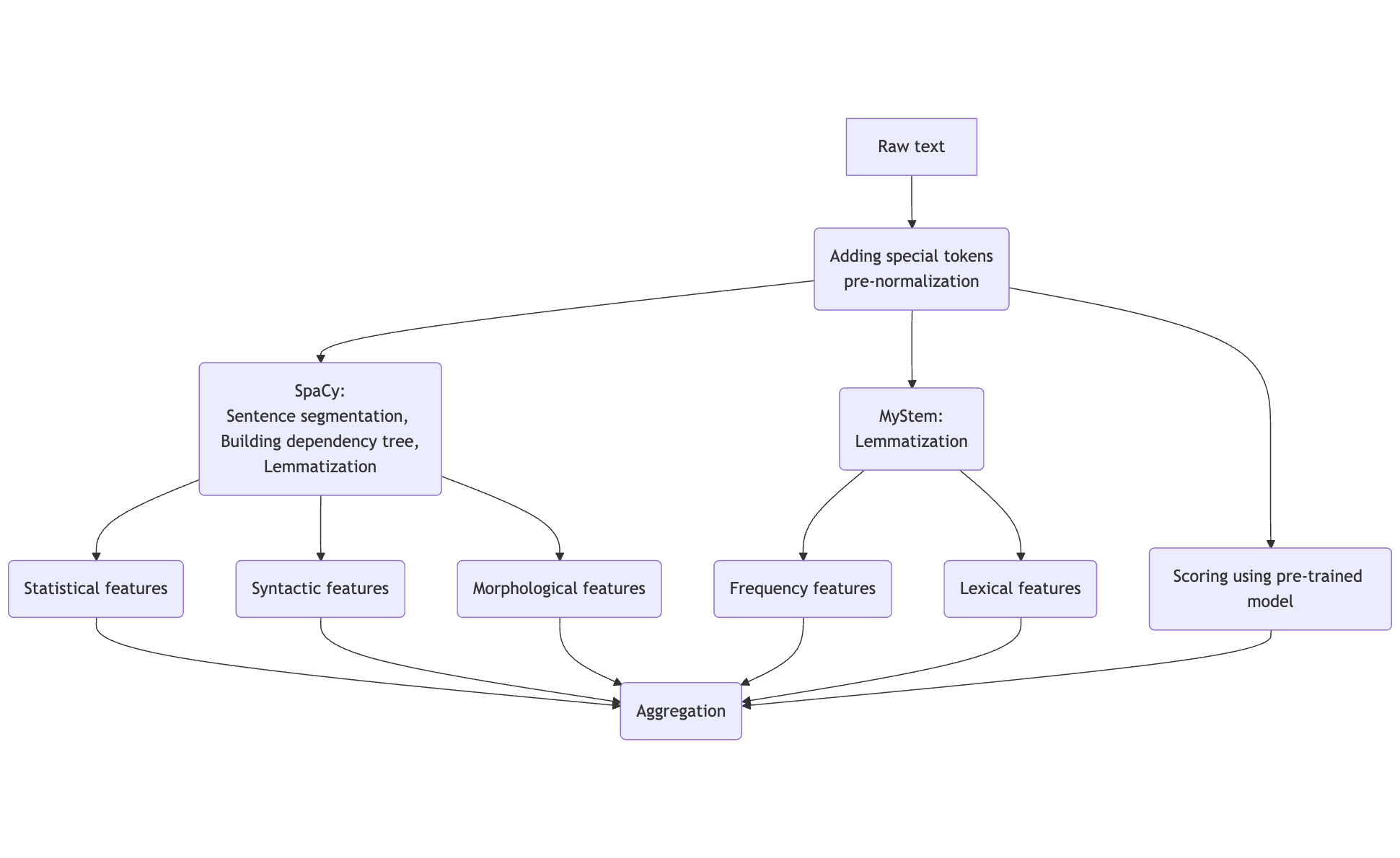}
\end{figure}

The platform processes texts from the submitted file sequentially. The analysis of each text in a sequence is carried out according to the pipeline presented in  Fig.~\ref{fig1}.

Thus, duplicate consecutive punctuation is removed and special tokens (e.g. emojis, numbers, links, etc.) are added. Then the given text is normalized and lemmatized using the selected approach depending on the analyzed group of categories. As a result, scores for the declared categories are calculated by the service backend. 

Texts are normalized for all categories, except categories that are calculated using pre-trained models. In these cases, raw texts with special tokens are fed directly into the model.

After aggregation, the scores for all 96 categories are recorded in the resulting files that can be downloaded by the user.

\section{Discussions}\label{sec14}

In the current study, we developed the service to automatically assess scores for the chosen categories considering the peculiarities of the Russian language. Although the service was built according to the widely used in the current scientific field methodology (LIWC), it has some limitations and open issues that need to be defined.

Some limitations relate to the result of applying the automated methods, despite their high overall efficiency. For instance, dependency parsing with SpaCy may generate an incorrect tree for a sentence if it contains improper punctuation, some punctuation marks are absent, or text has word order that can be difficult to analyze. As a result, the following computations may return inaccurate and unreliable scores.

Another factor that needs to be discussed is the content of lexical categories built for the internal dictionary. It may be limited and miss potentially suitable words and phrases denoting the corresponding categories. We collected content for categories on the basis of existing dictionaries, expanded it with corpus and automated methods. However, the semantic fields of the categories may still have incomplete coverage. The current version of the service does not include a full spectrum of linguistic features such as phrases, slang, emoticons, and idiomatic expressions that could enhance the category. We intend to make the dictionary more flexible in the future. 

However, it could be difficult to  resolve the fundamental issue with word count methods. As LIWC-based approaches use simple word analysis without considering the surrounding contexts for the target words, the computation could be inaccurate in a large corpus of texts. This aspect becomes significant especially for languages that have rich morphological system and use free word order. Such Russian texts as \textit{‘я никогда не любил тебя меньше, чем сейчас’} (\texttt{ I have never loved you less than I do now.}) and \textit{‘я невероятно люблю тебя, больше всего на свете’} (\texttt{I love you immensely, more than anything else in the world.}) represent the equal word percentages for the Love category, despite the general contexts of two texts are opposite.  

The word counting approach also ignores pragmatic and discoursal information, and communicative goals of the speaker. Some studies underline the importance of using tools other than direct word count,  including morphological, syntactic, and discursive definitions of the text. \cite{osipov2021method}. In more detail, the authors applied the Relational and Situational Analysis approach (“Mашина РСА”) method, which extracted almost 200 characteristics of Russian texts, including role structures. The given approach in comparison to LIWC underlines the deep structure of written texts. However, most categories in the study relate to linguistic aspects of texts, rather than to lexical and semantic representations that have already been noted in the field of psycholinguists. As a result, we could consider adding some approaches from the observed study to RusLICA, such as semantic parsing to improve the reliability of the service.

\section{Future work}\label{sec15}

We developed an analyzer that processes and evaluates Russian texts for observed categories considering the peculiarities of the language. The current version of the service was our first attempt at adapting LIWC and extending its categories to Russian. We focused on developing linguistic dimensions (syntactic and morphological) and collecting lexical entries for the resulting dictionary the service uses. 

However, the analyzer could be significantly improved in the future. To make the service more robust and efficient in providing more accurate results of evaluation, further studies should be aimed at a deeper understanding of the linguistic and psychological nuances of texts.
A primary direction for future development is extension of content for lexical categories. Enriching the internal dictionary with collocations, idiomatic expressions, emoticons, slang phrases would allow the analyzer to be applied in the increasing number of texts from various domains with specific vocabulary. The number of categories referring to emotional, social and biological representations should also be expanded using semantic dictionaries and automated methods, as they were used in the current study.  

Moreover, sentiment lexicon categorized by sources could be included in the category set in future studies. RuSentiLex \cite{loukachevitch2016creating} provides sentiment words and word collocations by fact, opinion and feeling sources with the analysis of ambiguity. However, including words from lexicon in the current version of the service was time- and resource-consuming, as the finalized set of words and collocations should be validated to avoid ambiguity.   

Another essential experimental component that has not been considered in the current study, but should be conducted in the future, is the evaluation of categories on the existing datasets. To evaluate the effectiveness and quality of created lexical categories, texts from Big-5 Personality or Well-being datasets can be scored and compared to the earlier studies \cite{panicheva2020matching}. This approach would help in further content extension and validation of the internal dictionary.

To make the service more robust and capable of analyzing text considering its context, more fine-tuned Language Models should be added to the service. The current version of the service uses one model to predict the emotion of texts. Extending the list of categories obtained by using predictions from LMs could become another crucial step. We expect integrating new LMs as an additional source of text analysis would cover more linguistic and psychological significant aspects.

\section{Conclusion}\label{sec16}

The current study introduces the first version of a free-to-use web service RusLICA, that analyzes Russian texts across psychologically significant and linguistic dimensions. Our approach enables automated analysis of linguistic characteristics (i.e., morphological and syntactic features) and lexical categories composed from existing semantic dictionaries, corpora, and thesauri. More precisely, the dictionary for lexical categories was built based on Russian semantic dictionaries, RuWordNet thesaurus, Russian National Corpus. Lexical categories were developed without relying on machine translation of the original LIWC dictionary. The current strategy for collecting lexical entries was designed to ensure that the internal dictionary can be conceptually and methodologically aligned with LIWC.

As a result, the service provides users both automatic assessment using parsers and LMs and word count analysis across selected lexical categories. It provides general statistical metrics, analyzes  syntax, morphology, word frequencies, word instances in lexical categories. In addition, the service can use fine-tuned language model for capturing context specificities of text.

\bmhead{Acknowledgements}

The author(s) declare that financial support was received for the research and/or publication of this article. The research was supported by RF state assignment 0138-2024-0020.

\bibliography{sn-bibliography}

@inproceedings{panicheva2020matching,
  title={Matching liwc with Russian thesauri: an exploratory study},
  author={Panicheva, Polina and Litvinova, Tatiana},
  booktitle={Conference on artificial intelligence and natural language},
  pages={181--195},
  year={2020},
  organization={Springer}
}

@article{pennebaker2015development,
  title={The development and psychometric properties of LIWC2015},
  author={Pennebaker, James W and Boyd, Ryan L and Jordan, Kayla and Blackburn, Kate},
 jounral={},
  year={2015}
}

@article{pennebaker1999linguistic,
  title={Linguistic styles: language use as an individual difference.},
  author={Pennebaker, James W and King, Laura A},
  journal={Journal of personality and social psychology},
  volume={77},
  number={6},
  pages={1296},
  year={1999},
  publisher={American Psychological Association}
}

@misc{pennebaker2001linguistic,
  title={Linguistic inquiry and word count: LIWC 2001},
  author={Pennebaker, James W},
  year={2001},
  publisher={Lawrence Erlbaum Associates}
}

@article{pennebaker2007linguistic,
  title={Linguistic inquiry and word count: LIWC [Computer software]},
  author={Pennebaker, James W and Booth, Roger J and Francis, Martha E},
  journal={Austin, TX: liwc. net},
  volume={135},
  year={2007}
}

@article{boyd2022development,
  title={The development and psychometric properties of LIWC-22},
  author={Boyd, Ryan L and Ashokkumar, Ashwini and Seraj, Sarah and Pennebaker, James W},
  journal={Austin, TX: University of Texas at Austin},
  volume={10},
  pages={1--47},
  year={2022}
}

@article{kailer2011russian,
  title={The russian liwc2007 dictionary},
  author={Kailer, Andreas and Chung, Cindy K},
  journal={Austin, TX: LIWC. net},
  year={2011}
}

@article{osipov2021method,
author = {Yenikolopov, S.N and Kuznetsova, YU.M and Osipov, G.S and Smirnov, I.V and Chudova, N.V},
year = {2021},
month = {12},
pages = {748-769},
title = {Metod relyatsionno-situatsionnogo analiza teksta v psikhologicheskikh issledovaniyakh},
volume = {18},
journal = {Psikhologiya. Zhurnal Vysshey shkoly ekonomiki},
doi = {10.17323/1813-8918-2021-4-748-769}
}

@unknown{davidchuk2023animore,
author = {Davidchuk, Nikita and Lubenec, I and Amentes, Artem},
year = {2023},
month = {03},
pages = {},
title = {ANIEMORE Otkrytaya biblioteka emotsiy v rechi cheloveka},
doi = {10.13140/RG.2.2.10999.80802}
}

@book{babenko2021emotions,
  title={Alfavit emotsiy: slovar'-tezaurus emotivnoy leksiki},
  author={Babenko, L{\^u}dmila Grigor'evna},
  year={2021}, 
  address={Ekaterinburg},
  publisher={Kabinetny{\u\i} ucheny{\u\i} }
}

@article{babenko1999verbs,
  title={Tolkovyy slovar' glagolov russkikh: ideograficheskoye opisaniye: angliyskiye ekvivalenty, sinonimy, antonimy},
  author={Babenko, L{\^u}dmila Grigor'evna},
  journal={},
  year={1999}
}

@misc{shvedova1998russkiy,
  title={Russkiy semanticheskiy slovar’. Tolkovyy slovar’, sistematizirovannyy po klassam slov i znacheniy. T. 1--4. Moscow: Azbukovnik},
  author={Shvedova, N Yu},
  year={1998},
  publisher={Russ}
}

@book{lyashevskaya2009dict,
  title={Chastotnyy slovar' sovremennogo russkogo yazyka po materialam yavleniy russkogo yazyka},
  author={Lyashevskaya, Ol'ga Nikolayevna and Sharov, Sergey Aleksandrovich},
  year={2009},
  address={Moscow},
  publisher={Obshchestvo s ogranichennoy otvetstvennost'yu" Izdatel'skiy tsentr "Azbukovnik"}
}

@article{duduau2021performing,
  title={Performing multilingual analysis with Linguistic Inquiry and Word Count 2015 (LIWC2015). An equivalence study of four languages},
  author={Dud{\u{a}}u, Diana Paula and Sava, Florin Alin},
  journal={Frontiers in Psychology},
  volume={12},
  pages={570568},
  year={2021},
  publisher={Frontiers Media SA}
}

@article{kern2016gaining,
  title={Gaining insights from social media language: Methodologies and challenges.},
  author={Kern, Margaret L and Park, Gregory and Eichstaedt, Johannes C and Schwartz, H Andrew and Sap, Maarten and Smith, Laura K and Ungar, Lyle H},
  journal={Psychological methods},
  volume={21},
  number={4},
  pages={507},
  year={2016},
  publisher={American Psychological Association}
}

@article{huston2019exploratory,
  title={Exploratory study of automated linguistic analysis for progress monitoring and outcome assessment},
  author={Huston, Jonathan and Meier, Scott and Faith, Myles and Reynolds, Amy},
  journal={Counselling and Psychotherapy Research},
  volume={19},
  number={3},
  pages={321--328},
  year={2019},
  publisher={Wiley Online Library}
}

@article{holtzman2019linguistic,
  title={Linguistic markers of grandiose narcissism: A LIWC analysis of 15 samples},
  author={Holtzman, Nicholas S and Tackman, Allison M and Carey, Angela L and Brucks, Melanie S and K{\"u}fner, Albrecht CP and Deters, Fenne Gro{\ss}e and Back, Mitja D and Donnellan, M Brent and Pennebaker, James W and Sherman, Ryne A and others},
  journal={Journal of Language and Social Psychology},
  volume={38},
  number={5-6},
  pages={773--786},
  year={2019},
  publisher={Sage Publications Sage CA: Los Angeles, CA}
}

@article{koutsoumpis2022kernel,
  title={The kernel of truth in text-based personality assessment: A meta-analysis of the relations between the Big Five and the Linguistic Inquiry and Word Count (LIWC).},
  author={Koutsoumpis, Antonis and Oostrom, Janneke K and Holtrop, Djurre and Van Breda, Ward and Ghassemi, Sina and de Vries, Reinout E},
  journal={Psychological Bulletin},
  volume={148},
  number={11-12},
  pages={843},
  year={2022},
  publisher={American Psychological Association}
}

@article{kwon2018will,
  title={How will we react to the discovery of extraterrestrial life?},
  author={Kwon, Jung Yul and Bercovici, Hannah L and Cunningham, Katja and Varnum, Michael EW},
  journal={Frontiers in Psychology},
  volume={8},
  pages={326082},
  year={2018},
  publisher={Frontiers}
}

@article{duduau2022development,
  title={The development and validation of the Romanian version of Linguistic Inquiry and Word Count 2015 (Ro-LIWC2015)},
  author={Dud{\u{a}}u, Diana Paula and Sava, Florin Alin},
  journal={Current Psychology},
  volume={41},
  number={6},
  pages={3597--3614},
  year={2022},
  publisher={Springer}
}

@inproceedings{van2017electronic,
  title={An electronic translation of the LIWC dictionary into Dutch},
  author={Van Wissen, Leon and Boot, Peter},
  booktitle={Electronic lexicography in the 21st century: Proceedings of eLex 2017 conference},
  pages={703--715},
  year={2017},
  organization={Lexical Computing}
}

@article{meier2019liwc,
  title={“LIWC auf Deutsch”: The development, psychometrics, and introduction of DE-LIWC2015},
  author={Meier, Tabea and Boyd, Ryan L and Pennebaker, James W and Mehl, Matthias R and Martin, Mike and Wolf, Markus and Horn, Andrea B},
  journal={PsyArXiv Preprints},
  number={uq8zt},
  year={2019},
  publisher={University of Zurich}
}

@article{bjekic2012razvoj,
  title={Razvoj srpske verzije re{\v{c}}nika za automatsku analizu teksta (LIWCser)},
  author={Bjeki{\'c}, Jovana and Eri{\'c}, Milica and Stojimirovi{\'c}, Elena and {\fontencoding{T1}\selectfont\DJ}oki{\'c}, Teodora and others},
  journal={Psiholo{\v{s}}ka istra{\v{z}}ivanja},
  volume={15},
  number={1},
  pages={85--110},
  year={2012},
  publisher={Univerzitet u Beogradu-Filozofski fakultet-Institut za psihologiju, Beograd}
}

@inproceedings{litvinova2018dynamics,
  title={Dynamics of an idiostyle of a Russian suicidal blogger},
  author={Litvinova, Tatiana and Litvinova, Olga and Seredin, Pavel},
  booktitle={Proceedings of the Fifth Workshop on Computational Linguistics and Clinical Psychology: From Keyboard to Clinic},
  pages={158--167},
  year={2018}
}

@misc{lukashevich2010thesauri,
  title={Tezaurusy v zadachakh informatsionnogo poiska},
  author={Lukashevich, Natal'ya Valentinovna},
  year={2010},
  publisher={M.: Izdatel'stvo MGU, 2011}
}

@inproceedings{loukachevitch2014ruthes,
  title={RuThes linguistic ontology vs. Russian wordnets},
  author={Loukachevitch, Natalia and Dobrov, Boris V},
  booktitle={Proceedings of the seventh global wordnet conference},
  pages={154--162},
  year={2014}
}

@inproceedings{loukachevitch2016creating,
  title={Creating a general Russian sentiment lexicon},
  author={Loukachevitch, Natalia and Levchik, Anatolii},
  booktitle={Proceedings of the Tenth International Conference on Language Resources and Evaluation (LREC'16)},
  pages={1171--1176},
  year={2016}
}

@inproceedings{vinogradova2017multi,
  title={Multi-level student essay feedback in a learner corpus},
  author={Vinogradova, OI and Lyashevskaya, ON and Panteleeva, IM},
  booktitle={Komp'yuternaya lingvistika i intellektual'nyye tekhnologii},
  pages={373--386},
  year={2017}
}

\newpage

\begin{appendices}

\section{Extracted linguistic categories}\label{allcategories}
\footnotesize 
\setlength{\tabcolsep}{5pt}
\begin{xltabular}{\textwidth}{@{} L  C @{}}
\toprule%
\textbf{Category} & \textbf{Abbreviations} \\
\midrule
\midrule
\multicolumn{2}{c}{General Statistics} \\
\midrule
\midrule
Number of words & N\_words\\
Average number of words per sentence & mean\_Nwords\_sent\\
Maximum number of words per sentence & max\_Nwords\_sent\\
Average word length  & mean\_len\_words\\
Maximum word length & max\_len\_words\\
\\
Number of repeated words (the sum of the occurrences of repeated words is calculated regardless of the lemma) & N\_repeated\_words\\
\\
Number of occurrences of the most repeated word (calculated regardless of the lemma) & N\_most\_repeated\_word\\
Number of emojis&N\_emogies \\
Number of hashtags& N\_hashtags\\
Number of numerical tokens &N\_nums \\
Number of links&N\_urls \\
\midrule
\midrule
\multicolumn{2}{c}{Frequency Categories} \\
\midrule
\midrule
Average IPM frequency & mean\_ipm\\
Average D frequency & mean\_d\\
Minimum IPM frequency & min\_freq\_ipm \\
Minimum D frequency&min\_freq\_d\\

\midrule
\midrule
\multicolumn{2}{c}{Syntactic Categories} \\
\midrule
\midrule
Average syntactic tree depth (across sentences in the text) & mean\_sent\_depth\\
\\
Maximum syntactic tree depth (across sentences in the text) & max\_synt\_depth\\
\\
Minimum syntactic tree depth (across sentences in the text) & min\_synt\_depth\\
\\
Number of relative clause modifiers of a nominal & N\_acl\:relcl\\
Number of adnominal clauses& N\_acl\\
Number of adverbial modifiers &N\_advcl \\
Total number of relative and subordinate clauses described above & N\_relative\\
Number of conjuncts& N\_conj\\
Number of discourse element& N\_discourse\\
Number of subordinate conjunctions& N\_sconj\\
Number of coordinating conjunctions& N\_cconj\\

\midrule
\midrule
\multicolumn{2}{c}{Morphological Categories} \\
\midrule
\midrule

Number of first-person pronouns&N\_pronn\_pers\_first \\
Number of second-person pronouns&N\_pronn\_pers\_sec \\
Number of third-person pronouns&N\_pronn\_pers\_third \\
Number of adverbs&N\_adv \\
Number of adjectives&N\_adj \\
Number of comparative adjectives&N\_adj\_degree\_compar \\
Number of superlative adjectives &N\_adj\_degree\_super \\
Number of sentence-root verbs in the indicative mood&N\_verb\_root\_mood\_ind \\
Number of sentence-root verbs in the imperative mood&N\_verb\_root\_mood\_imp \\
Number of sentence-root verbs in the imperfective aspect&N\_verb\_root\_aspect\_imp \\
Number of sentence-root verbs in the perfective aspect&N\_verb\_root\_aspect\_perf \\
Number of sentence-root verbs in the active voice&N\_verb\_root\_voice\_act\\ 
Number of sentence-root verbs in the passive voice&N\_verb\_root\_voice\_pass\\
Number of sentence-root verbs in the middle voice&N\_verb\_root\_voice\_mid\\
Number of sentence-root verbs in the past tense&N\_verb\_root\_tense\_past\\
Number of sentence-root verbs in the present tense&N\_verb\_root\_tense\_pres\\
Number of sentence-root verbs in the future tense&N\_verb\_root\_tense\_fut\\
Number of verbs in the indicative mood&N\_verb\_mood\_ind\\
Number of verbs in the imperative mood&N\_verb\_mood\_imp\\
Number of verbs in the imperfective aspect&N\_verb\_aspect\_imp\\
Number of verbs in perfective aspect&N\_verb\_aspect\_perf\\
Number of verbs in the active voice&N\_verb\_voice\_act\\
Number of verbs in the passive voice&N\_verb\_voice\_pass\\
Number of verbs in the middle voice&N\_verb\_voice\_mid\\
Number of verbs in the past tense&N\_verb\_tense\_past\\
Number of verbs in the present tense&N\_verb\_tense\_pres\\
Number of verbs in the future tense&N\_verb\_tense\_fut\\
\midrule
\midrule
\multicolumn{2}{c}{Classification Categories (based on pre-trained LMs)} \\
\midrule
\midrule

predicted emotion category (using  \textit{Aniemore/rubert-tiny2-russian-emotion-detection} model)& emotion\_prediction \\

\botrule

\end{xltabular}

\section{Lexical categories}\label{secA1}

\footnotesize 
\renewcommand{\arraystretch}{0.9} 
\setlength{\tabcolsep}{5pt}
\begin{xltabular}{\textwidth}{@{} L L C C @{}}
\toprule%
\textbf{Category} & \textbf{Abbreviations} & \textbf{Word examples} & \textbf{Word Count} \\
\midrule
\midrule
\multicolumn{4}{c}{I. Linguistic Dimensions} \\
\midrule
\midrule
Modality  & модальность & \textit{истинность, неотвратимость, неправда, обман, условность} &  158 \\
\\
Negations  & отрицания & \textit{ничто, никогда, ничуть, незачем, никак}  & 41 \\
\\
Intensifiers  & интенсификатор & \textit{абсолютный, донельзя, колоссально, наповал, отчаянно, очень}  & 245 \\
\midrule
\midrule
\multicolumn{4}{c}{II. Psychological processes} \\
\midrule
\midrule
\multicolumn{4}{l}{2.1 Affective processes} \\
\midrule
\multicolumn{4}{l}{\hspace{0.2cm}2.1.1 Positive emotions} \\
\\
\hspace{0.7cm}Happiness&счастье&\textit{счастливчик, счастливый, удовольствие, эйфория, праздник}&130 \\
\\
\hspace{0.7cm}Joy&радость&\textit{радостно, жизнерадостный, оптимизм, повеселеть, весело}&214 \\
\\
\hspace{0.7cm}Love&любовь&\textit{влюбленность, любить, нравиться, обожание, фаворит, флирт}&243 \\
\\
\hspace{0.7cm}Inspiration&вдохновение&\textit{вдохновлять, увлекающийся, страсть, окрыленность, муза}&62 \\
\multicolumn{4}{l}{\hspace{0.2cm}2.1.2 Negative emotions} \\
\\
\hspace{0.7cm}Grief&горе&\textit{беда, выплакивать, гибельный, горесть, злосчастие, мученический}&222 \\
\\
\hspace{0.7cm}Sadness&грусть&\textit{грустный, апатичный, изнывать, пессимизм, понурый}&222 \\
\\
\hspace{0.7cm}Anxiety&беспокойство&\textit{беспокоиться, взвинчиваться, нервничать, тревога, тревожность}&157 \\
\\
\hspace{0.7cm}Anger&злость&\textit{гнев, злиться, зловещий, злость, ожесточенный}&172 \\
\\
\hspace{0.7cm}Fear&страх&\textit{боязнь, жуткий, испуг, малодушный, панический}&179 \\
\\
\hspace{0.7cm}Loneliness&одиночество&\textit{одинокий, покинутость, нелюдимость, одиночка, отшельник}&109 \\
\\
\hspace{0.7cm}Shame&стыд&\textit{позор, постыдиться, стыдить, угрызение, сконфузить}&160 \\
\\
\hspace{0.4cm}Swear words & ругательства & \textit{блядь, мудак, пиздец, херня} & 191 \\
\midrule
\multicolumn{4}{l}{2.2 Social processes} \\
\midrule
\hspace{0.4cm}Feud & вражда & \textit{рознь, скандал, ругаться, ссориться, стравливать} & 90 \\
\\
\hspace{0.4cm}Arrogance & высокомерие & \textit{наглый, снисходительный, надменно, высокомерно, беспардонный} & 37 \\
\\
\hspace{0.4cm}Resentment & обида  & \textit{обижать, обижаться, досадно, дуться, унижение} & 85 \\
\\
\hspace{0.4cm}Interpersonal relationships & межличностные отношения & \textit{мирить, невзлюбить, опекать, осуждение, поддерживать} & 269  \\
\\
\hspace{0.4cm}Communication & речевая деятельность & 
\textit{беседа, бормотать, говорить, наорать, просьба }& 360  \\
\\
\hspace{0.4cm}Social relations &  социальные отношения & \textit{воспитывать, вразумлять, изгнание, лидерство, опекать} & 388  \\
\multicolumn{4}{l}{\hspace{0.2cm}2.2.1 Social referents} \\
\\
\hspace{0.7cm}Male referents&мужские референты&\textit{мужчина, друг, муж, юноша, мальчик, житель}& 93\\
\\
\hspace{0.7cm}Female referents&женские референты&\textit{женщина, мама, мисс, подруга, суженая, союзница}& 169 \\
\\
\hspace{0.7cm}Family&семья&\textit{брат, близнец, брачный, замужество, поколение, родительство}& 207\\
\\
\hspace{0.7cm}Friendship&дружба&\textit{дружба, единомышленник, побратим, подруга, приятель}& 87 \\

\midrule
\multicolumn{4}{l}{2.3 Physical} \\
\midrule
\hspace{0.4cm}Illness & здоровье (болезни) & \textit{инфекция, нерв, насморк, ожог, перитонит, сердцебиение} & 283 \\
\\
\hspace{0.4cm}Mental health &  ментальное здоровье & \textit{аддикция, апатичный, ипохондрия, паранойя, психоз, тревожный} & 419 \\
\\
\hspace{0.4cm}Sexual & сексуальность & \textit{интим, либидо, секс, эротизм, похоть} & 247 \\
\\
\hspace{0.4cm}Death & окончание существования (смерть) & \textit{гибель, казнь, кремирование, летальный, смерть, умерший} & 179 \\
\midrule
\multicolumn{4}{l}{2.4 Perception} \\
\midrule
\hspace{0.4cm}Auditory & слышать & \textit{аудио, грохот, звучать, слух, слышать, трезвонить} & 224 \\
\\
\hspace{0.4cm}Visual & видеть & \textit{видеть, глядеть, зрение, осмотр, тень, тусклый} & 138 \\
\\
\hspace{0.4cm}Motion & движение & \textit{забег, залезать, подбегать, погоня, походка, бегун} & 571 \\
\midrule
\multicolumn{4}{l}{2.5 Motives} \\
\midrule
\hspace{0.4cm}Curiosity & любопытство & \textit{интересно, любознательный, увлекаться, любопытство, интриговать} & 32 \\
\\
\hspace{0.4cm}Passion & влечение & \textit{симпатичный, притягательный, одержимый, манящий, азарт, увлечение} & 151 \\
\midrule
\multicolumn{4}{l}{2.6 Cognition} \\
\midrule
\hspace{0.4cm}Belief & вера & \textit{верить, доверие, уверенный, убедительный, преданность} & 63 \\
\\
\hspace{0.4cm}Cognition and Mindset & сознание. образ мыслей & \textit{гипотеза, довод, допущение, кредо, мечта, мозг, осознание} & 353 \\
\multicolumn{4}{l}{\hspace{0.2cm}2.6.1 Cognitive processes} \\
\\
\hspace{0.7cm}Causation & причинность & \textit{вследствие, мотив, потому, ввиду, первопричина} & 31 \\
\\
\hspace{0.7cm}Doubt & сомнение & \textit{скептик, сомневаться, неуверенный, нерешительность, возможно} & 43 \\
\\
\hspace{0.7cm}Certitude & уверенность & \textit{безусловно, истина, корректный, отчетливость, подлинность} & 90 \\
\midrule
\midrule
\multicolumn{4}{c}{I I I.   Lifestyle} \\
\midrule
\midrule
\multicolumn{4}{l}{3.1 Metaphysical processes} \\
\midrule
\hspace{0.4cm}Religion & религия & \textit{верование, грех, душа, оберег, церковь, шаманство} & 248 \\
\midrule
\midrule
\multicolumn{4}{c}{I V.  Time and Space orientation} \\
\midrule
\midrule
Time & время & \textit{век, временной, дата, декабрь, минута} & 309 \\
\\
Space & пространство & \textit{высота, диаметр, метро, север, угол} & 638 \\
\botrule

\end{xltabular}

\end{appendices}

\end{document}